\documentclass[letterpaper, 10 pt, conference]{ieeeconf}  

\IEEEoverridecommandlockouts                              

\usepackage{graphicx}
\usepackage{xcolor}
\usepackage{multirow}
\usepackage{graphicx}
\usepackage{algorithm}
\usepackage{algpseudocode}
\usepackage{amsmath}
\usepackage{array}
\usepackage{changepage} 
\usepackage{comment} 
\usepackage{graphicx} 
\usepackage{float} 

\usepackage{balance}

\newcommand{\papername}{STREAK}
\newcommand{\lowerbound}{finetuned\_GRAPH}
\newcommand{\upperbound}{complete\_GRAPH}

\title{\LARGE \bf
STREAK: Streaming Network for Continual Learning of Object Relocations under Household Context Drifts
}

\author{Ermanno Bartoli$^{1}$, Fethiye Irmak Doğan$^{1}$ and Iolanda Leite$^{1}$
\thanks{$^{1}$Ermanno Bartoli, Fethiye Irmak Doğan and Iolanda Leite are with Faculty of Robotics Perception and Learning,
        KTH Royal Institute of Technology, Stockholm, Sweden}
        }

\begin{document}

\maketitle
\thispagestyle{empty}
\pagestyle{empty}

\begin{abstract}
In real-world settings, robots are expected to assist humans across diverse tasks and still continuously adapt to dynamic changes over time. For example, in domestic environments, robots can proactively help users by fetching needed objects based on learned routines, which they infer by observing how objects move over time. However, data from these interactions are inherently non-independent and non-identically distributed (non-i.i.d.), e.g., a robot assisting multiple users may encounter varying data distributions as individuals follow distinct habits. This creates a challenge: integrating new knowledge without catastrophic forgetting.
To address this, we propose \papername{} (Spatio Temporal RElocation with Adaptive Knowledge retention), a continual learning framework for real-world robotic learning. It leverages a streaming graph neural network with regularization and rehearsal techniques to mitigate context drifts while retaining past knowledge. Our method is time- and memory-efficient, enabling long-term learning without retraining on all past data, which becomes infeasible as data grows in real-world interactions. We evaluate \papername{} on the task of incrementally predicting human routines over 50+ days across different households. Results show that it effectively prevents catastrophic forgetting while maintaining generalization, making it a scalable solution for long-term human-robot interactions.

\end{abstract}

\section{INTRODUCTION}
\label{sec:introduction}

Robots deployed in domestic environments can be expected to assist multiple users with diverse routines. This requires the robot to continuously adapt as it interacts with different users following diverse routines, leading to context drift in the robot's learning processes. 
Such real-world data is non-independent and non-identically distributed (non-i.i.d.), making it difficult for robots to generalize across environments~\cite{fini2020online}. Moreover, privacy concerns, memory limitations, and time constraints make it impractical to store and retrain on all past data as new interactions accumulate~\cite{LESORT202052}. In such cases, robots should develop mechanisms to integrate new knowledge efficiently while retaining previously acquired information and avoiding catastrophic forgetting \cite{Kirkpatrick_2017}.

Prior work in proactive assistance~\cite{patel2022proactive, 9889357,9494681, 9524521}, human action prediction~\cite{10.3233/AIS-200556, 8542683, 8307470}, and healthcare~\cite{WILSON2019258,9134708} assumes static, predefined knowledge, making these approaches unsuitable for real-world scenarios where robots must learn from new users and changing routines. Without mechanisms to mitigate catastrophic forgetting \cite{Kirkpatrick_2017}, robots risk losing previously acquired behaviors. 
Addressing this, Continual Learning (CL) offers a solution by enabling robots to learn incrementally without discarding prior knowledge \cite{shaheen2022continual}. 
Inspired by its diverse applications in computer vision research~\cite{shaheen2022continual}, CL has been explored for robotics~\cite{shaheen2022continual} and human-robot interaction~\cite{knowledgeacquisitionkge, 9223564} settings. 
Still, these methods often assume well-structured task boundaries or overlook real-world constraints such as memory and time limitations~\cite{9223564}, making them difficult to apply in dynamic, real-world environments.
Addressing this, we suggest a novel CL framework that handles context drifts and ensures long-term knowledge retention for adaptive and scalable robotic assistance in dynamic households.

\label{sec:robot_demo}
\begin{figure}[t!]
\centering
\begin{minipage}{0.5\textwidth}
  \centering
  \includegraphics[width=0.8\linewidth]{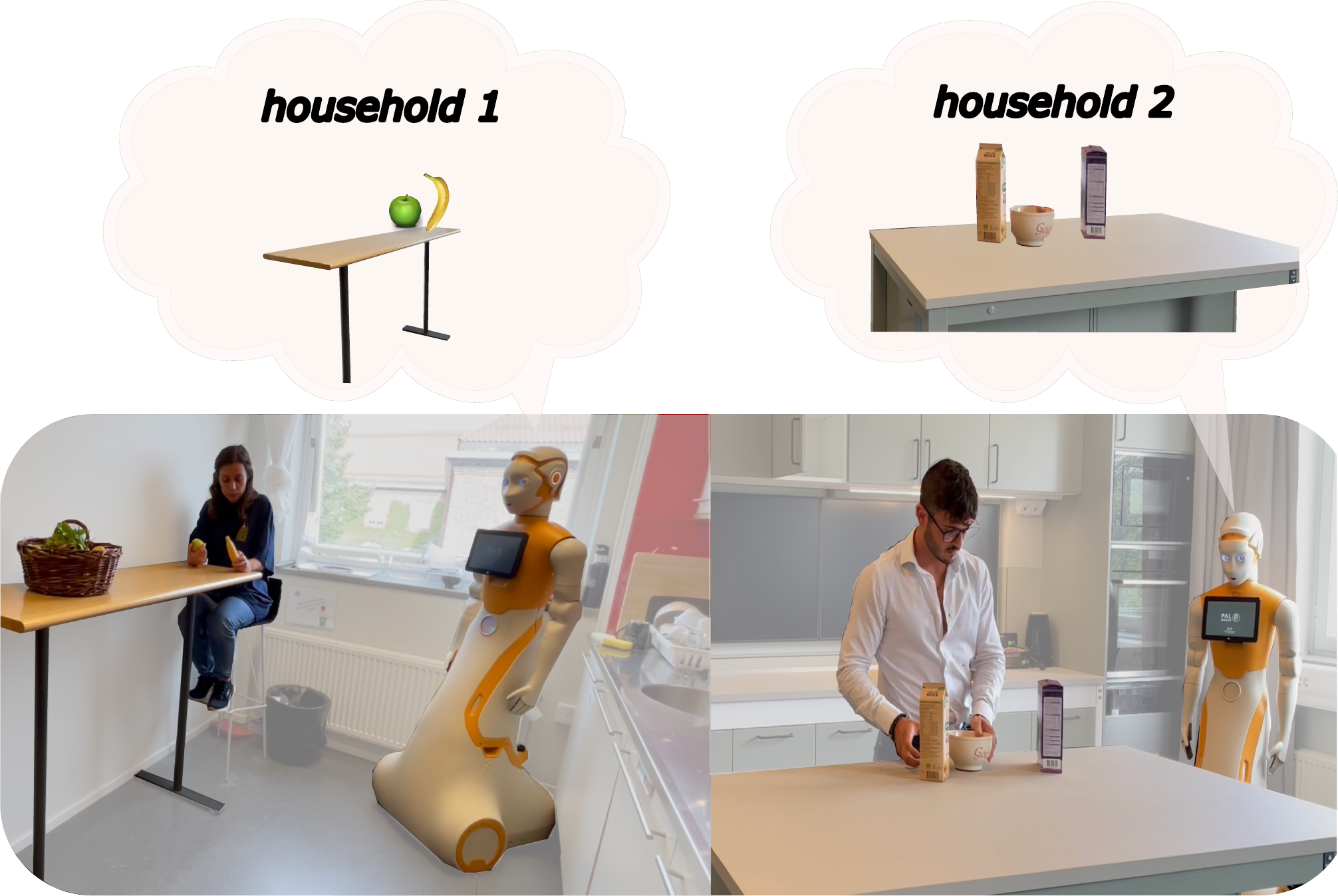}
    \caption{The robot inspects the spatio-temporal dynamics of the objects\\ in two different households.}
    \label{fig:robot_demo}
\end{minipage}%
\end{figure}

In this paper, we propose \papername{} (\textbf{S}patio \textbf{T}emporal \textbf{RE}location with \textbf{A}daptive \textbf{K}nowledge retention), a CL framework for proactive robot assistance leveraging a streaming graph neural network to learn human routines over time and across different households. The robot observes patterns of multiple humans interacting with objects in their environment, continuously adapting as it encounters new users and homes.
This is achieved through a streaming graph neural network that integrates regularization in the loss function with a rehearsal method, ensuring that the most important past experiences are retained and replayed. 
To assess its effectiveness, we compared \papername{} with a generative graph neural network used in \cite{patel2022proactive}, which considers a static version of this problem where the model is trained independently in each environment. 
Experimental results demonstrate that \papername{} effectively mitigates catastrophic forgetting when sequentially exposed to new households while maintaining accurate predictions in previously seen environments. Additionally, our approach is significantly more 
time- and memory-efficient 
and robustly incorporates new tasks from unseen households compared to the baseline method. 
Finally, we demonstrated the use case of our method in a real-world scenario with the ARI robot.

\section{RELATED WORK}
\label{sec:related_works}

\begin{figure*}[t!]
\centering
\begin{minipage}{0.99\textwidth}
  \centering
  \includegraphics[width=\linewidth]{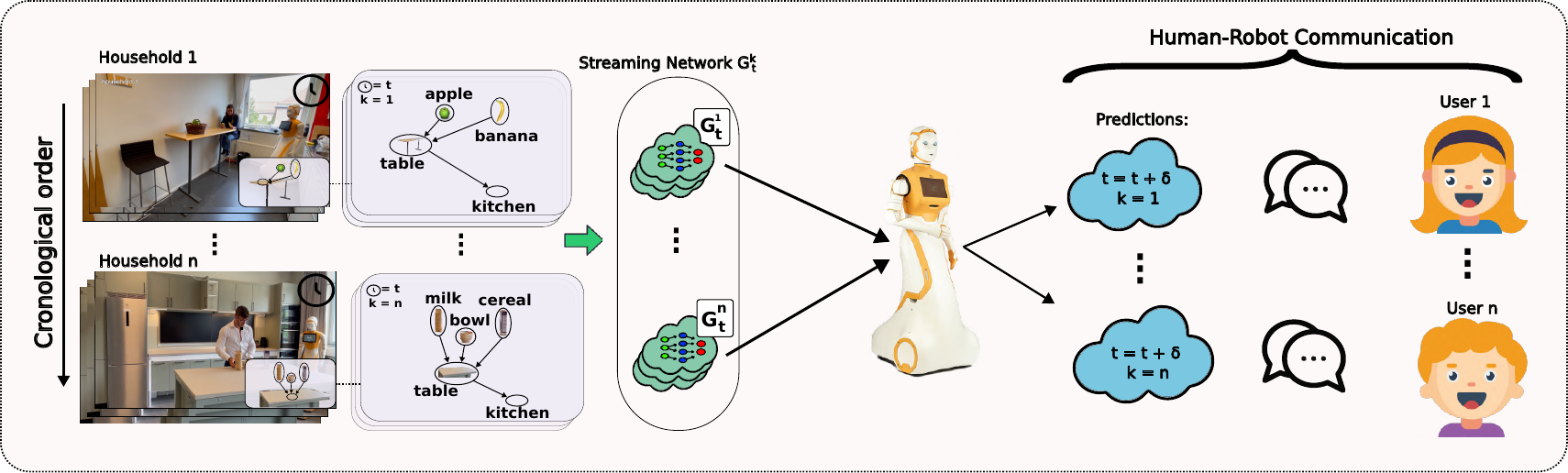}
  \caption{Overview of proposed \papername{} framework. The robot acquires the graph state through user action observation, learning each household in sequence. For each household, the robot assembles the respective graph state, $G^{k}_{t}$, at a given time step $t$. Finally, it predicts dynamic spatial object relocations according to user routines for each household.}
  \label{fig:overview}
\end{minipage}%
\end{figure*}

\textit{Proactive robot assistance} involves the development of robots capable of assisting users without being explicitly queried, enabling them to actively engage with their environment and anticipate user needs~\cite{WILSON2019258,9134708}. While initial research primarily focused on collaborative setups where humans and robots explicitly worked together~\cite{8542683, 8307470}, more recent work has explored proactive scenarios. In these setups, robots autonomously predict user actions or requirements and provide assistance without interrupting user workflows~\cite{9494681, 9524521, 10.3233/AIS-200556}.
For instance,~\cite{10.3233/AIS-200556} introduced an action graph in a kitchen environment to predict user actions through observation while ensuring no disruption to their routine. Similarly,~\cite{qian2022environment} addressed spatial–temporal coordination in human–robot collaboration by leveraging demonstrations. Another approach utilized Graph Neural Networks (GNNs) to analyze object movements~\cite{patel2022proactive}, allowing the robot to predict and assist with object relocation tasks in daily routines.
These works represent significant progress in enabling robots to assist proactively by anticipating user needs. However, in real-world scenarios, effective proactive assistance requires robots to continuously adapt to dynamic environments, various users and changing user behaviors.

\textit{Continual Learning (CL)} contributes to lifelong robot learning by enabling adaptation to changing data distributions over time~\cite{LESORT202052}. Initial research applied incremental learning to context modeling in robotics~\cite{8462925, 8593633}, and later works explored the integration of CL with reinforcement learning for robot navigation~\cite{kalifou2019continual, traore2019discorl, 9345478}. More recently, CL has been leveraged to assess the social appropriateness of robot actions using Bayesian Networks~\cite{10.3389/frobt.2022.669420}, while other studies have incrementally and hierarchically constructed Boltzmann Machines to learn novel scene contexts over time~\cite{8462925}. Despite these advancements, CL remains challenging for assistive robots, which assimilate diverse information from real-world environments where data distributions change over time~\cite {8911341}.

To mitigate catastrophic forgetting when continually learning, various strategies have been explored~\cite{LESORT202052}. Dynamic architectures that evolve over time have been proposed~\cite{8462925, 8593633, rusu2016progressive, knowledgeacquisitionkge, parisi2017lifelong}. Regularization approaches, including drop-out~\cite{goodfellow2013empirical}, early stopping~\cite{MALTONI201956}, and advanced constraint-based methods~\cite{sharif2014cnn, lee2017overcoming, fernando2017pathnet}, have also been widely used. Additionally, rehearsal-based techniques~\cite{churamani2020clifer, 8100070} store samples from previous tasks to preserve knowledge while learning new ones~\cite{aljundi2019gradient}. While effective, these approaches require balancing memory constraints and computational efficiency.

Graph Neural Networks (GNNs) also suffer from catastrophic forgetting when trained incrementally~\cite{yuan2023continual}. To counter this, prior studies have introduced experience replay~\cite{zhou2021overcoming}, gradient-based sample selection~\cite{aljundi2019gradient}, and transformations that treat graph nodes as independent graphs~\cite{wang2022lifelong}. 

To the best of our knowledge, ours is the first study considering proactive robot assistance combined with incremental robot learning. Building on prior work in proactive robot assistance, we employ a combination of regularization techniques and rehearsal-based learning, preserving previously encountered samples using the Mean Feature Criteria~\cite{zhou2021overcoming} while detecting new patterns to retain past knowledge~\cite{wang2020streaming}. Our approach extends CL for proactive assistance, formulating object relocation as a Streaming Neural Network problem to integrate CL in adapting to novel environments, such as different homes and users, ensuring more adaptive and effective robotic behavior~\cite{9223564}.

\section{BACKGROUND}
\label{sec:background}
\textit{\textbf{Streaming Neural Network.}} The concept of a Streaming Network has been introduced in~\cite{wang2020streaming}, where they denoted as $G = (G^1, G^2, \ldots, G^K)$, wherein each 
\begin{equation}
    G^k = G^{k-1} + \Delta G^k
    \label{eq:streaming_network_def}
\end{equation}    
 symbolizes an attributed graph corresponding to task $k$, and $\Delta G^k$, is the changes of node attributes and network structures for the task $k$. Subsequently, the authors expanded upon this foundation to define Streaming Graph Neural Networks (Streaming GNNs), an evolution of conventional GNNs tailored for a streaming context. In this model, given the streaming network $G$, the objective is to determine a sequence of optimal parameter sets $(\theta^1, \theta^2, \ldots, \theta^K)$, with each $\theta^k$ representing the optimal parameters for the GNNs associated with task $k$. A recommended approach for training the streaming network involves specifically training each $G^k$ on $\Delta G^k$ by utilizing $\theta^{k-1}$ as the initialization point. Nonetheless, should $\Delta G^k$ induce alterations in the patterns previously recognized by $\theta^{k-1}$ within $G^{k-1}$, the risk of catastrophic forgetting emerges. To evade the potential deterioration in the representation of nodes and edges within $G^{k-1}$, it is fundamental to implement CL strategies, thereby preserving the model's ability to maintain and update its knowledge base effectively.

\textit{\textbf{Spatio-Temporal Object Dynamics Model.}} The concept of the spatio-temporal object dynamics model has been delineated in~\cite{patel2022proactive} in an endeavor to comprehend the movements of objects over time. Patel et al. conceptualized the environment using a graph notation $G_t = \{V_t, E_t\}$, which encapsulates the state of the graph at time $t$. Here, $G_t$ is characterized as an in-tree, with nodes ${v_{i}^{k}} \in V$ symbolizing objects $o_{i}$ and their respective locations $l_{i}$. Additionally, the edges ${e_{i,j}} \in E$ extend from every node barring the root. The primary goal, given the graph state $G_{t}$, is to accurately forecast the consequent graph state at a future time step $\delta$, denoted as $\hat{G}_{t+\delta}$. This model sets the foundation for predicting the relocation of objects within a predefined temporal scope, thereby facilitating a deeper understanding of their dynamic behavior in spatial and temporal dimensions. However, the dynamic and evolving nature of environments, where the same objects can be relocated differently across households or even within the same household by different users, requires continual adaptation to capture these varying patterns of object location changes.

\section{METHODOLOGY: \papername{}}
\label{sec:problem formulation}
\textbf{Task Description.} The task involves predicting object relocations in dynamic household environments, where the robot must infer how objects move between locations (e.g., a mug moving from a table to a kitchen sink), reflecting user routines as these object location changes are caused by user actions. 
By modelling these changes, our goal is to enable assistive robots to learn and adapt incrementally over time, ensuring they generalize across multiple users and environments while retaining knowledge of past interactions. The dataset for this task consists of graph-based representations, with nodes representing objects and locations and edges encoding the relationship "is-in". 

We propose two main components. Firstly, we extend the Streaming Neural Network formulation to model spatio-temporal dynamics of object displacements to continually learn over time. Secondly, to mitigate catastrophic forgetting, we adopt two CL strategies: the introduction of a penalty to the loss function to ensure controlled changes in the Streaming Neural Network when context drifts happen (see Sec.\ref{subsec:sptsnn}), and the inclusion of a dynamically allocated memory buffer that keeps the most significant former information (see Sec. \ref{subsec:method}).

\subsection{SPATIO-TEMPORAL STREAMING NEURAL NETWORK}
\label{subsec:sptsnn}
We learn object location changes in dynamic environments through a Spatio-Temporal Streaming Neural Network, where context shifts over time. The streaming network is defined similarly as in eq.~\ref{eq:streaming_network_def}, where each graph $G^k$ evolves from the previous one by incorporating new changes $\Delta G^k$.
\noindent Differently, we define: 
\begin{equation}
    \Delta G^k = \sum_{m=1}^T G^k_m - G^k_{m-1},
\end{equation}
which encapsulates all the temporal evolution within the network for each task, from $m=1$ to $m=T$.
This formulation captures evolving contexts as new tasks are introduced. However, as the number of tasks grows, explicitly summing all past changes becomes intractable. Instead, following \cite{wang2020streaming}, we approximate updates at time $t$ using:
\begin{equation}
\Delta G^k_t = G^{k}_{t} - G^{k}_{t-1}.
\label{eq:new_delta}
\end{equation}
This allows us to model global context shifts based only on recent graph states. We use the predicted future graph $\hat{G}^{k}_{t+\delta}$ to infer object relocations caused by human actions, where each relocation $r(o_i, l_1, l_2)$ represents an object $o_i$ moving from location $l_1$ to $l_2$.

The model learns $\Phi(G^{k}_{t}) \longrightarrow p(G^{0:k}_{t+\delta})$ to predict future graph states while preserving knowledge from previous tasks. The ultimate goal is to optimize $(\theta^1, \theta^2, ... , \theta^K)$, where $\theta^k$ represents the optimal parameters for task $k$, ensuring generalization across past tasks $[0:k-1]$. 

\subsection{OVERCOMING CATASTROPHIC FORGETTING}
\label{subsec:method}

When the GNN encounters a context drift due to a new task to be learned, it is crucial to consolidate previously learned patterns to prevent catastrophic forgetting. We tackle this by introducing an additional term in the loss function, which constrains the variation of the model parameters to remain close to the optimal values learned during the previous task $k-1$. This consolidation loss penalizes large deviations from the previous task’s parameters, thereby preventing the model from focusing solely on the new task and forgetting prior knowledge. Additionally, we prioritize the simplicity and efficiency of our approach to ensure real-time functionality on a physical robot. Hence, the loss function is formulated as follows:

\begin{equation}
L^{k} = L^{k}_{\text{model}} + L^{k}_{\text{consolidation}},
\label{eq:loss_equation}
\end{equation}
\begin{equation}
L^{k}_{\text{consolidation}} = \frac{\lambda}{2} \sum_{i} \textbf{F}_i (\theta_i^{k} - \theta_{i}^{k-1})^2.
\end{equation}

Eq.~(\ref{eq:loss_equation}) combines two loss terms, namely $L^{k}_{\text{model}}$ and $L^{k}_{\text{consolidation}}$, to guide the learning process.
The term $L^{k}_{\text{model}}$ represents the model loss introduced in~\cite{patel2022proactive}. It's the combination of: $L_{\text{class}}$ cross-entropy loss for node classification, $L_{\text{location}}$ cross-entropy loss for edge, and $L_{\text{context}}$ cosine embedding loss that enforces consistency in the context representation.
We compose it with $L^{k}_{\text{consolidation}}$, which computes the deviation between the current model's parameters ($\theta^k_i$) and the optimal values of the parameters obtained from the previous task ($\theta_{i}^{k-1}$). $\textbf{F}_i$ is the component of the Fisher Information Matrix for the $i$-th parameter, and it indicates the importance. The deviation is squared and multiplied by a weight factor $\lambda$, thus incorporating objectives that promote the preservation of learned patterns. 

In order to ensure efficiency, we minimize the amount of data stored in memory by discarding non-informative samples. Given the problem of context drifting, we identify the most informative data as the closest to the average feature vector, as suggested in~\cite{wang2020streaming}. Accordingly, we compute the average embedded feature vector $c_l$ as follows:

\begin{equation}
c_l=\frac{1}{\left|V^k\right|+\left|E^k\right|+\left|C^k\right|} \sum_{\substack{v_i \in V^k \\ e_i \in E^k \\ c_i \in C^k}} \boldsymbol{h_i^v}+\boldsymbol{h_i^e}+\boldsymbol{h_i^c},
\end{equation}%
\noindent where $c_l$ is the sum of individual embedding feature vectors $\boldsymbol{h_i^v}$, $\boldsymbol{h_i^e}$, and $\boldsymbol{h_i^c}$ associated with nodes, edges, and time encoding, respectively, and dividing it by the total number of elements in sets $V^k$, $E^k$, and $C^k$. 
In this formulation, $V^{k}$ is the set of training nodes, $E^{k}$ is the set of edges, and $C^{K}$ is the set of time encoding for task $k$. 

%
%

In order to ensure sustainability over time and prevent memory overload when dealing with multiple tasks, we retain only the most informative data. To achieve this, we devise an approach that constructs a Memory Buffer $\mathcal{M}_{k}$ for each learning session, which is defined as follows:
\begin{equation}
\mathcal{M}_{k}=\sum_{j=1}^k \frac{1}{\beta \cdot (k-j+1)} D_j,
\label{eq:memory_buffer}
\end{equation}%
\noindent where $D_j$ represents the dataset at session $j$. The sum iterates over previous sessions from $j=1$ to $k$. The term $\frac{1}{\beta \cdot (k-j+1)}$ represents the weight assigned to the dataset at session $k$, indicating how the influence of previous datasets gradually decreases as we move further.
This buffer contains the current dataset whose distribution describes the current task, as well as selected experiences from the past. The selection process involves dynamically adjusting the number of samples in the previous memory based on their informativeness.

This approach allows us to strike a balance between memory efficiency and the preservation of valuable knowledge from previous sessions. By adaptively controlling the number of samples in the memory buffer $\mathcal{M}_{k}$, we can effectively manage the storage requirements while retaining the most informative data for CL.
The choice of $\beta$ determines the trade-off between memory efficiency and the preservation of previously learned knowledge (its efficiency analyzed in Section~\ref{time_memory_analyze} in detail). 
Algorithm~\ref{alg:continual_training} shows the overall training procedure of \papername{}.

\begin{algorithm}
\caption{\papername{} training pipeline}
\begin{algorithmic}[1] 
    \State Initialize the nodes $V^{0}_{0}$, edges $E^{0}_{0}$, time encoding $C^{0}_{0}$
    \State  $G^{0}_{0} = \{V^{0}_{0}, E^{0}_{0}\}$
    \For{task $k$ $\in$ K}
    \For{time step $t$ $\in$ T}
        \State $L^{k}_{\text{consolidation}} = \frac{\lambda}{2} \sum_{i} \textbf{F}_i (\theta_i^{k} - \theta_{i}^{k-1})^2$
      \State $L^{k} = L^{k}_{\text{model}} + L^{k}_{\text{consolidation}}$ \Comment{Compute loss}
        \State $\Delta G^k_t = G^k_t - G^k_{t-1}$
        \State $G^{k}_{t} = G^{k-1}_{t} + \Delta G^{k}_{t}$ \Comment{Graph update}
      \EndFor
      \State $c_l=\frac{1}{\left|V^k\right|+\left|E^k\right|+\left|C^k\right|} \sum_{\substack{v_i \in V^k \\ e_i \in E^k \\ c_i \in C^k}} \boldsymbol{h_i^v}+\boldsymbol{h_i^e}+\boldsymbol{h_i^c}$
      \State $\mathcal{M}_{k}=\sum_{j=1}^k \frac{1}{\beta \cdot (k-j+1)} D_j$ \Comment{Buffer update}
    \EndFor
\end{algorithmic}
\label{alg:continual_training}
\end{algorithm}


\section{EVALUATION}
\label{sec:Evaluation}

We evaluate our model on mitigating catastrophic forgetting when retaining previous knowledge. In addition, we want to maintain the satisfactory predictive performance of new tasks, as it should not be solely focused on knowledge retention while potentially sacrificing its predictive abilities on upcoming data. Furthermore, as CL has been noted to be time and memory-efficient~\cite{LESORT202052}, particularly in applications involving real robots that interact with humans~\cite{ayub2024interactive}, we also evaluate these components. Therefore, our evaluation encompasses the following aspects: knowledge retention, predictive performance on new tasks, and time/memory efficiency.\\

\textit{\textbf{Dataset.}} We used the HOMER dataset introduced in~\cite{patel2022proactive}. This dataset consists of a collection of regular activities recorded from various individuals over a span of several weeks. These activities were drawn from five distinct households, 
over a comprehensive duration of 60 days. We partitioned the dataset into two segments: a training set spanning 50 days and a test set covering the remaining 10 days.

Since the routines come from five different household environments, where users have different routines, a natural context shift occurs when we sequentially consider the data from each household. As a result, the data inherently introduces context drifts, eliminating the need for additional preprocessing to simulate them.
As the dataset does not include multiple users within a single household, each household corresponds to a unique user. Consequently, considering different households is equivalent to modeling sequential interactions with distinct users. This makes the dataset a valid and relevant scenario for studying context drift, as the challenges of learning from separate households mirror those of interacting with different users over time.
Finally, the recorded behaviors are transformed into a graph-based representation, where nodes represent either objects or locations, and edges indicate the presence or absence of relation "is-in" between the nodes. This graph structure, denoted as $G^{k}_{t}$, serves as input for the model.

\textit{\textbf{Metrics.}} Given the task of predicting object relocations based on human activities, we categorize the predictions into distinct outcomes. We separate objects that were used by humans during the interval $[t:t + \delta]$ from those that remained unused during the same period.

For objects that were used by humans, predictions are categorized as follows: objects correctly predicted to have been moved to their correct locations are labeled as \textbf{"Moved Correct"}; objects correctly predicted to have been moved but to the wrong locations are labeled as \textbf{"Moved Wrong"}; and objects that were moved but were wrongly predicted as not having been moved are labeled as \textbf{"Moved Missed"}.

For unused objects (i.e., those whose final and original locations remain the same), predictions are categorized into two outcomes: objects correctly predicted as not having been moved are labeled as \textbf{"Unmoved Correct"}, while objects incorrectly predicted as having been moved to a different location are labeled as \textbf{"Unmoved Wrong"}.\\

\textit{\textbf{Benchmarks.}} The benchmarks were established through the definition of both lower and upper bounds, against which our model's performance was compared. 
The lower bound (\textbf{\lowerbound{}}) was obtained by sequentially fine-tuning the model described in~\cite{patel2022proactive} across the datasets. It is important to note that the lower bound obtained from \lowerbound{} is not a chance level but is the result of an SOTA model from the non-incremental approach of \cite{patel2022proactive}.

To establish the upper bound (\textbf{\upperbound{}}), the GTM model of \cite{patel2022proactive} was jointly trained with shuffled data from all preceding tasks. This approach is common practice in CL settings to identify maximum performance limits, as highlighted in~\cite{Aljundi_2019_CVPR}.  The \upperbound{} model might not always be applicable in real-world scenarios, as robots typically cannot access all data from different environments simultaneously due to potential issues related to unpredictable and 
dynamic nature of such environments, along with growing data size and resource limitations. Both the lower and upper bounds used the optimal set of hyperparameters identified in~\cite{patel2022proactive}.
\\ 

\textit{\textbf{Knowledge Retention}.} To evaluate \textit{knowledge retention}, we trained \papername{}, \lowerbound{}, and \upperbound{} on all datasets sequentially. After each training phase, we tested the models on all previously encountered datasets to assess their ability to retain knowledge over time. Specifically, for each learning session $LS_k$, the models were trained 
incrementally on the current dataset $D_k$ (\papername{}), fine-tuned on the current dataset $D_k$ (\lowerbound{}), and trained on the joint dataset $[D_0:D_k]$ (\upperbound{}). They were then evaluated on all datasets separately up to $D_k$. 
Table~\ref{table:prove_cf} illustrates this process for \lowerbound{}, where each row represents a learning session. The table also demonstrates the phenomenon of catastrophic forgetting, as performance on previously encountered datasets degrades progressively with each additional learning session, indicating a loss of prior knowledge and motivating the need for a continual learning approach.\\

\begin{table}[t!]
\centering
\resizebox{0.49\textwidth}{!}{%
\begin{tabular}{ |c||c|c|c|c|c|c| } 
\cline{3-7}
\multicolumn{1}{c}{} & \multicolumn{1}{c|}{} & \multicolumn{5}{c|}{\textbf{Test}} \\
\cline{3-7}
\multicolumn{1}{c}{} & & D0 & D1 & D2 & D3 & D4 \\
\hline
\multirow{5}{*}{\rotatebox[origin=c]{90}{\centering\textbf{Train}}} & $\lowerbound{}_{0}$ on D0 & 44.35 & - & - & - & -\\ 
& $\lowerbound{}_{1}$ on D1 & 8.69 & 36.98 & - & - & -\\ 
& $\lowerbound{}_{2}$ on D2 & 21.8 & 6.74 & 44.33 & - & -\\ 
& $\lowerbound{}_{3}$ on D3 & 20.24 & 9.38 & 13.75 & 34.11 & - \\ 
& $\lowerbound{}_{4}$ on D4 & 12.39 & 13.12 & 23.5 & 2.43 & 35.88\\ 
\hline
\end{tabular}}
\caption{\small Each row of the table shows the performance of the \lowerbound{} finetuned up to $D_{k}$ on the test datasets $D_{0:k}$.}
\vspace{-1.5em}
\label{table:prove_cf}
\end{table}

\textit{\textbf{Prediction on the new task}}. This evaluation seeks to assess the model's ability to not only retain previously learned knowledge but also make accurate predictions for new, upcoming tasks. To ensure a fair comparison when evaluating our approach’s performance on new tasks, we selected the upper bound \textbf{\upperbound{}} as a benchmark. This allows us to assess our model’s performance relative to one that has access to all data from all tasks, thereby assessing the 
capabilities of our proposed approach.
For evaluating the \textit{prediction on new tasks}, at each learning session $LS_k$, the models were trained incrementally on the current dataset $D_k$ (\papername{}) and on the joint dataset $[D_0:D_k]$ (\upperbound{}). Then, instead of evaluating the models on all datasets $D_{0:k}$, we focused on evaluating their performance solely on the last dataset encountered $D_k$, which corresponds to the most recent task learned by the model.\\

\textit{\textbf{Time and Memory Efficiency}}. 
In addition to evaluating the models' performance, we also assessed their time and memory efficiency. To do so, we compared the training and inference times as well as the memory requirements of \papername{} against \upperbound{}, which serves as the upper bound and has the highest performance. This comparison provided insights into the computational costs of our approach and its scalability in real-world scenarios. Furthermore, we analyzed the memory usage of the two models by considering the number of samples required during the training process. This allowed us to assess the trade-offs between model complexity and computational resources, offering a quantitative evaluation for practical implementation.

For the time and memory complexity analysis, we considered the scenario in which, at each learning session $LS_k$, a new dataset $D_k$ related to the new task is introduced. We compared two models: the \upperbound{} model, which retains all previously encountered datasets and trains on the entire dataset history $[D_0:D_k]$, and the \papername{}, which trains only on the current dataset $D_k$ combined with the memory buffer $\mathcal{M}_k$.\\

\textit{\textbf{Implementation Details.}} 
We experimented with different hyperparameter values: $\lambda \in {80, 100, 200}$ and $\beta \in {5, 10, 15}$. The models were trained for \{25, 50, 100\} epochs using a batch size of 1 to simulate online learning and tested for proactivity by varying the prediction horizon $\delta$, which represents the time window for anticipating future object relocations. Specifically, the model was tested to predict object movements at $t + \delta$, where $\delta$  ranged from 10 minutes to 120 minutes, with intervals of 30 minutes.  ReLU was used as the activation function, and optimization was performed with Adam at a learning rate of $10^{-3}$.

For clarity, we present the results based on the best set of parameters identified during the experimentation phase. The optimal configuration was found to be $[\lambda = 200, \beta = 10]$, with the model being trained for 50 epochs. This combination was determined to yield the best performance, balancing the trade-offs between computational efficiency and model accuracy. We show results for $\delta = 10$ minutes.
All experiments were conducted on a desktop system equipped with an Intel RTX 3080 GPU with 10GB of dedicated memory, an 11th Gen Intel(R) Core(TM) i7-11700K @ 3.60GHz CPU, and 32GB of RAM. 

\section{RESULTS}
\label{sec:result}

\subsection{Knowledge retention.} 
Table \ref{table:ourresult} presents the performance of \papername{} in comparison with the \lowerbound{} and \upperbound{} approaches. \papername{} demonstrated superior performance in knowledge retention compared to the \lowerbound{} as shown in the last row of Table \ref{table:ourresult}. \papername{} achieved higher accuracy in predicting both moved and unmoved objects.
Specifically, \papername{} achieved a mean average of 25.2\% correct predictions on moved objects, outperforming the \lowerbound{}'s average of 17.46\%.
Moreover, we observe that \papername{} performs closely to the \upperbound{} in terms of accuracy on moved objects for each task. While the \upperbound{} model has the advantage of being trained on the entire dataset from the start, \papername{}, with its CL approach, can achieve performance comparable to the \upperbound{}.  The emphasis on correctly moved objects is due to its greater significance compared to other metrics: In practice, a user would prefer the robot to miss an object that should be moved rather than incorrectly moving an object, as the latter requires a more complex recovery operation to retrieve the object from an unknown location.

The results indicate that \papername{} performs effectively in an incremental learning setting, coming close to the performance of an \upperbound{} model that has the advantage of full access to the entire dataset. This showcases the efficacy of \papername{} in handling sequential data and its ability to strike a balance between knowledge retention and predictive performance, making it a valuable solution for CL scenarios.

\begin{figure}[t!]
\centering
\begin{minipage}{0.5\textwidth}
  \centering
  \includegraphics[width=\linewidth]{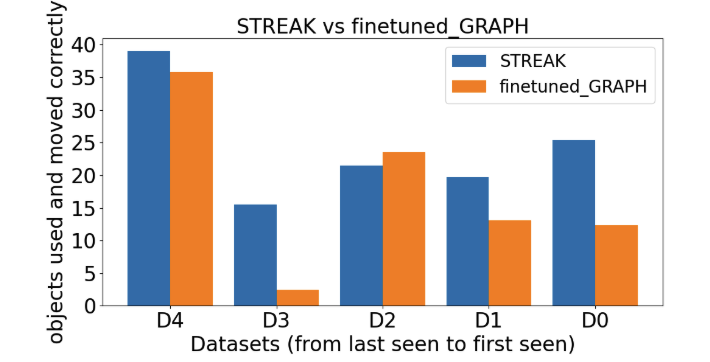}
    \caption{Evaluation of "Moved Correct" of \lowerbound{} and \papername{} on the 5 datasets, after the models have been trained sequentially on $D_0 \rightarrow D_4$}
    \label{fig:1v1}
\end{minipage}%
\end{figure}

We conducted an additional experiment to demonstrate the performance decline of the \lowerbound{} model compared to the more stable performance of the \papername{} model when both underwent the same sequence of datasets (${D_0 \rightarrow D_1 \rightarrow D_2 \rightarrow D_3 \rightarrow D_4}$). The results, shown in Figure \ref{fig:1v1}, present the accuracy of correctly moved objects for simplicity. 
The accuracy of \papername{} on correctly moved objects, while experiencing some inevitable loss in prediction skills, demonstrates a consistently higher and more stable trend compared to \lowerbound{}. This indicates that \papername{} has a better ability to maintain accuracy and make reliable predictions across tasks.

\begin{table}[t!]
    \centering
    \renewcommand{\arraystretch}{1.1} 
    \resizebox{0.5\textwidth}{!}{%
        \begin{tabular}{|l|c|c|c|c|c|}
            \hline
            \multicolumn{1}{|c|}{} & \multicolumn{3}{c|}{\% Moved Objects} & \multicolumn{2}{c|}{\% Unmoved Objects} \\
            \cline{2-6}
            \multicolumn{1}{|c|}{\textbf{}} & Correct & Wrong & Missed & Correct & Wrong \\
            \hline
            \lowerbound{} (lower bound) & 17.46 & 4.25 & 78.28 & 98.51 & 1.49\\ \hline
            \upperbound{} (upper bound) & 28.15 & 3.36 & 68.49 & 99.74 & 0.26 \\
            \hline
            \textbf{\papername{} (ours)} & 25.20 & 4.43 & 70.08 & 98.58 & 1.42 \\
            \hline
        \end{tabular}}
    \caption{Performance of \papername{} (ours) with respect to \lowerbound{} (Lower Bound), and \upperbound{} (Upper Bound).}
    \label{table:ourresult}
    \vspace*{-1\baselineskip}
\end{table}

\begin{table}[t!]
    \centering
    \renewcommand{\arraystretch}{1.1} 
    \resizebox{\linewidth}{!}{%
        \begin{tabular}{|l|c|c|c|c|c|}
            \hline
            & $D_0$ & $D_1$ & $D_2$ & $D_3$ & $D_4$ \\
            \hline
            \upperbound{} & 34.91 & 25.44 & 21.89 & \textbf{26.08} & 32.45 \\
            \hline
            \textbf{\papername{}} & \textbf{35.71} & \textbf{27.73} & \textbf{23.42} & 20.72 & \textbf{39.03} \\
            \hline
        \end{tabular}}  
    \caption{Comparison between the accuracy of the correctly moved objects on the last seen datasets included incrementally.}
    \label{table:comparisonlastdataset}
    \vspace*{-1\baselineskip}
\end{table}

\subsection{Prediction on new tasks}
Table \ref{table:comparisonlastdataset} shows the results of this evaluation. \papername{}, when compared to \upperbound{}, demonstrated better predictive capabilities for the new upcoming tasks, with the only exception of D3. 
We attribute the improved performance of \papername{} to its further focus on the current task during training. Unlike models trained on the entire dataset, where data from all tasks are mixed, incremental learning allows the model to concentrate on the most recent task, potentially improving its ability to incorporate new information. However, in \papername{}, this advantage is balanced by regularization techniques that retain knowledge from previous tasks
to ensure the trade-off between learning new tasks and retaining past knowledge.
\begin{figure}[t!]
\centering
\begin{minipage}{0.5\textwidth}
  \centering
  \includegraphics[width=1.0\linewidth]{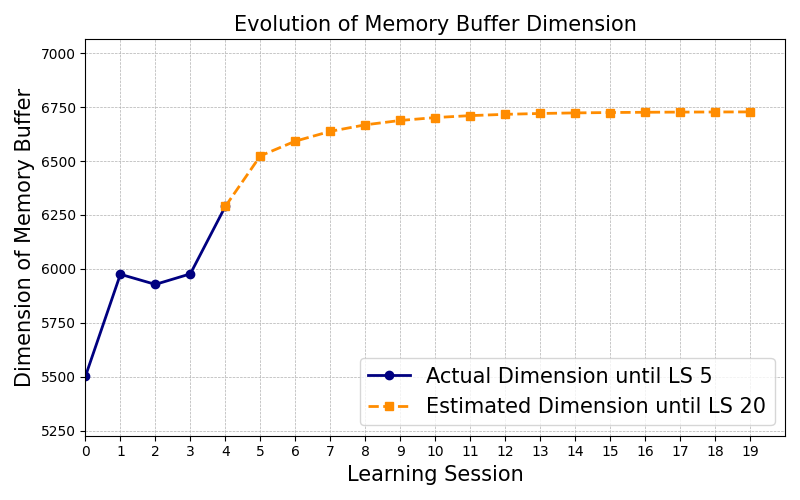}
    \caption{The dimension (number of samples) of $\mathcal{M}_{k}$ across the five learning sessions (blue line), and the estimated size of $\mathcal{M}_{k}$ after 10 learning sessions (orange line) where we considered all the datasets of the same size, equal to the mean of the five existing ones.}
    \label{fig:trend_data_bounded}
\end{minipage}%
\end{figure}
\label{time_memory_analyze}

\subsection{Time and memory efficiency}
To evaluate the practical viability of our approach, we analyzed the time and memory requirements of both models trained on the same sequence of datasets (see Table \ref{table:time} and Table \ref{table:memory}). \papername{} demonstrates efficient time and memory usage, making it a more practical choice for real-world scenarios. In contrast, the \upperbound{} model, while yielding slightly better results, exhibits disproportionate growth in time and memory requirements, making it intractable for long-term use.

Previously, we demonstrated that \papername{} outperforms \lowerbound{}, achieving results comparable to \upperbound{}. However, the \upperbound{} model’s reliance on full data access, which enhances performance, is often impractical as all data may not be available in real-world scenarios. Furthermore, \upperbound{} models tend to have high memory and time demands, especially in long-term scenarios involving multiple datasets. In contrast, CL can manage these constraints effectively.

To support our claims, we examined the evolution of the Memory Buffer dimension $\mathcal{M}{k}$ in Figure \ref{fig:trend_data_bounded} across five learning sessions. This analysis provides insights into how the buffer size evolves as new tasks are encountered. Additionally, we predicted the buffer size for another 15 learning sessions, assuming that each new dataset has a constant size equal to the mean of the existing five datasets. This forecast demonstrates that the dimension of $\mathcal{M}{k}$ remains bounded, ensuring efficiency and preventing excessive growth.

\begin{table}[t!]
    \centering
    \resizebox{\linewidth}{!}{
    \begin{tabular}{|l|c|c|c|c|c|c|}
        \hline
         & \multicolumn{6}{c|}{\textbf{Time requirement} (lower is better)}\\
        \cline{2-7}
        & LS0 & LS1 & LS2 & LS3 & LS4 & Total \\
        \hline
        \textbf{\upperbound{}} & 22.5 & 42.5 & 65.8 & 70.0 & 108.0 & 308.8\\
        \textbf{\papername{}} & \textbf{21.6} & \textbf{33.3} & \textbf{35.8} & \textbf{35.8} & \textbf{40} & \textbf{166.5}\\
        \hline
    \end{tabular}
    }
    \caption{Training time (in mins) required for each learning session.}
    \label{table:time}
\end{table}
\begin{table}[t!]
    \centering
    \resizebox{\linewidth}{!}{
        \begin{tabular}{|l|c|c|c|c|c|c|}
            \hline
             & \multicolumn{6}{c|}{\textbf{Memory requirement (lower is better)}}\\
            \cline{2-7}
            & LS0 & LS1 & LS2 & LS3 & LS4 & Total\\
            \hline
            \textbf{\upperbound{}} & 5175 & 10350 & 15420 & 20700 & 25875 & 77520\\
            \textbf{\papername{}} & 5175 & \textbf{5693} & \textbf{6038} & \textbf{6268} & \textbf{6421} & \textbf{29595}\\
            \hline
        \end{tabular}
        }
        \caption{Memory requirement (number of samples) during training for each learning session.}
        \label{table:memory}
        \vspace*{-1\baselineskip}
\end{table}

\section{ROBOT DEMONSTRATION}

We implemented a proof-of-concept robotics demonstration using scenarios derived from the breakfast routines of two distinct households, labeled Household 1 and Household 2. This demonstration was executed in a kitchen environment using the ARI robot, as depicted in Figure \ref{fig:robot_demo}. This demonstration allowed us to present the use case of our approach without requiring additional training data, as we used the model trained on the existing routines. The data acquisition occurred sequentially, with the robot transitioning from household 1 to household 2. We then tasked the robot with proactive prediction, anticipating the timing and content of breakfast. 
As shown in the supplementary material video, our approach enables the robot to predict object replacements correctly and provide adequate assistance to the user even in formerly learned households, as opposed to non-continual methods, which suffer from catastrophic forgetting. Given the physical limitations of the robot, which prevent it from carrying objects, the assistance is provided verbally, with the robot informing the user about the objects they will need. However, this limitation is specific to the robot itself; with a robot capable of carrying objects, the approach could enable the robot to fetch the items directly, eliminating the need for verbal communication.

\section{Considerations and Limitations}
The results suggest practical considerations for real-world applications. In structured settings like education, if all tasks are known in advance, \upperbound{} can be used, as continual adaptation is less critical. In contrast, in scenarios where the robot primarily focuses on domain adaptation and performance on the final task, 
\lowerbound{} may be preferred.
However, in situations where the robot must both adapt to new tasks and retain prior knowledge, such as in healthcare where a robot assisting patients (e.g., delivering medications) has to learn new patients’ needs without forgetting those of previous ones, \papername{} offers a compelling solution. By balancing efficiency and long-term retention, \papername{} achieves performance close to \upperbound{} while remaining computationally feasible.


Our framework includes some limitations. For instance, retaining the knowledge completely is still a challenge due to the inherent complexity of the problem as the dataset exhibits highly diverse distributions.
Future work could explore better techniques to mitigate catastrophic forgetting, handle concept drift, and enhance adaptability to evolving data. As the number of households or users grows, prioritizing memory and time efficiency may come at the cost of knowledge retention.
Finally, exploring techniques for dynamically adjusting the network architecture to adjust to changing task requirements or data distributions could potentially further improve the adaptation capabilities.

\section{CONCLUSION}
\label{sec:conclusion}

In this paper, we propose a novel approach for incremental learning in the context of detecting object relocation. We achieve this by introducing a novel CL framework using a streaming graph neural network designed to learn spatio-temporal object relocations. We ensure the retention of previously acquired knowledge using regularization and rehearsal techniques. The experimental results demonstrate the effectiveness of our approach in achieving accurate object relocation detection under household context drifts. Our method demonstrates improved knowledge retention capabilities, proving to be efficient in terms of both memory and time efficiency. 
Overall, \papername{}  demonstrates promising performance in incremental learning for real-world scenarios, making a step forward in the deployment of autonomous robots in human environments.




\bibliographystyle{unsrt}
\balance{\bibliography{example}}

\end{document}